\documentclass{article}
\usepackage{ijcai18}

\usepackage{times}
\usepackage{xcolor}
\usepackage{soul}
\usepackage[utf8]{inputenc}
\usepackage[small]{caption}

\usepackage{url}

\usepackage[ruled]{algorithm2e}
\usepackage[caption=false]{subfig}
\usepackage{amssymb}
\usepackage[cmex10]{amsmath}
\usepackage{booktabs}
\usepackage[figuresright]{rotating}

\usepackage{epsfig}
\usepackage{graphicx}
\usepackage{amsmath}
\usepackage{amssymb}
\usepackage{multirow}
\usepackage{comment}

\title{Unifying and Merging Well-trained Deep Neural Networks for Inference Stage}

\author{
	Yi-Min Chou$^{1,2}$,
	Yi-Ming Chan$^{1,2}$,
	Jia-Hong Lee$^{1,2}$,
	Chih-Yi Chiu$^3$,
	Chu-Song Chen$^{1,2}$
	\\
	$^1$ Institute of Information Science, Academia Sinica, Nankang, Taipei, Taiwan \\
	$^2$ MOST Joint Research Center for AI Technology and All Vista Healthcare\\
	$^3$ National Chiayi University,    Chiayi City, Taiwan  \\
	\tt\small \{chou, yiming, honghenry.lee, song\}@iis.sinica.edu.tw,
	\tt\small chihyi.chiu@gmail.com,
}

\begin{document}
	
	\maketitle
	
	\begin{abstract}
		We propose a novel method to merge convolutional neural-nets for the inference stage.
		Given two well-trained networks that may have different architectures that handle different tasks, our method aligns the layers of the original networks and merges them into a unified model by sharing the representative codes of weights.
		The shared weights are further re-trained to fine-tune the performance of the merged model.
		The proposed method effectively produces a compact model that may run original tasks simultaneously on resource-limited devices.
		As it preserves the general architectures and leverages the co-used weights of well-trained networks, a substantial training overhead can be reduced to shorten the system development time.
		Experimental results demonstrate a satisfactory performance and validate the effectiveness of the method.
	\end{abstract}
	
	\section{Introduction}
	
	The research on deep neural networks has gotten a rapid progress and achievement recently.
	It is successfully applied in a wide range of artificial intelligence (AI) applications, including computer vision, speech processing, natural language processing, bioinformatics, etc.
	To handle various tasks, we usually design different network models and train them with particular datasets separately, so they can behave well for specific purposes.
	However, in practical AI applications, it is common to handle multiple tasks simultaneously, leading to a high demand for the computation resource in both training and inference stages.
	Therefore, how to effectively integrate multiple network models in a system is a crucial problem towards successful AI applications.
	
	This paper tackles the problems of merging multiple well-trained (known-weights) feed-forward networks and unifying them into a single but compact one.
	The original networks, whose architectures may not be identical, can be either single or multiple source input.
	After unification, the merged network should be capable of handling the original tasks but is more condensed than the whole original models.
	Our approach (NeuralMerger) contains two phases:
	
	\noindent \textbf{Alignment and encoding phase}: First, we align the architectures of neural network models and encode the weights
	such that they are shared among the networks.
	The purpose is to unify the weights so that the filters and weights of different neural networks can be co-used.
	
	\noindent \textbf{Fine-tuning phase}: Second, we fine-tune the merged model with partial or all training data (calibration data). A method following the concept of distilling dark knowledge of neural networks in \cite{HintonDistilling14} is employed in this phase.
	
	\begin{figure} [t]
		\begin{center}
			\includegraphics[width=0.75\columnwidth]{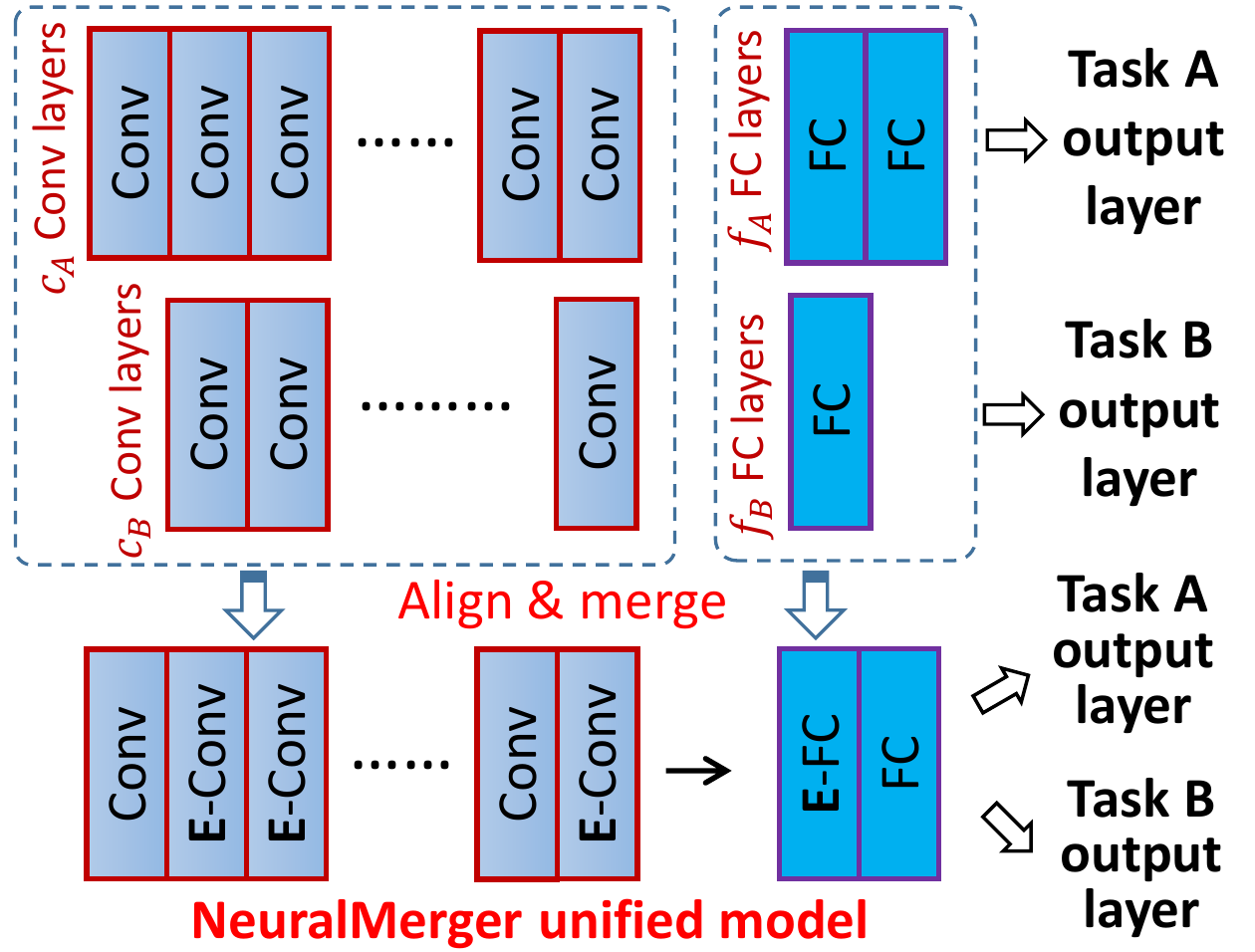}
		\end{center}
		\vspace{-0.1cm}
		\caption{Two tasks accomplished by feed-forward networks, where model A (or B) consists of $c_A$ (or $c_B$) convolution and $f_A$ (or $f_B$) fully-connected layers, respectively. Our NeuralMerger unifies the two models into a single one consisting of $max(c_A, c_B)$ convolution and $max(f_A, f_B)$ fully-connected layers for the model inference; \textbf{E}-Conv and \textbf{E}-FC are referred to as the \emph{Jointly-\textbf{E}ncoded} \emph{convolution} and \emph{fully-connected} layers, respectively.
		}
		\label{fig:concept}
	\end{figure}
	
	Neural network models may have very different topologies.
	Currently, this study focuses on merging feed-forward networks, while merging networks with loops remains a future work.
	A modern feed-forward network consists of several kinds of layers, including convolution, pooling, and full-connection, which is generally referred to as a convolutional network (CNN).
	When merging two CNNs, our approach aligns the same-type layers (convolution; full-connection) into pairs.
	The layers in a pair are merged into a single layer that shares a common weight codebook through the proposed encoding scheme.
	The codebooks in the merged single model can be further trained via back-propagation algorithm; it thus can be fine-tuned to seek for performance improvement.
	
	\begin{figure} [t]
		\begin{center}
			\includegraphics[width=\columnwidth]{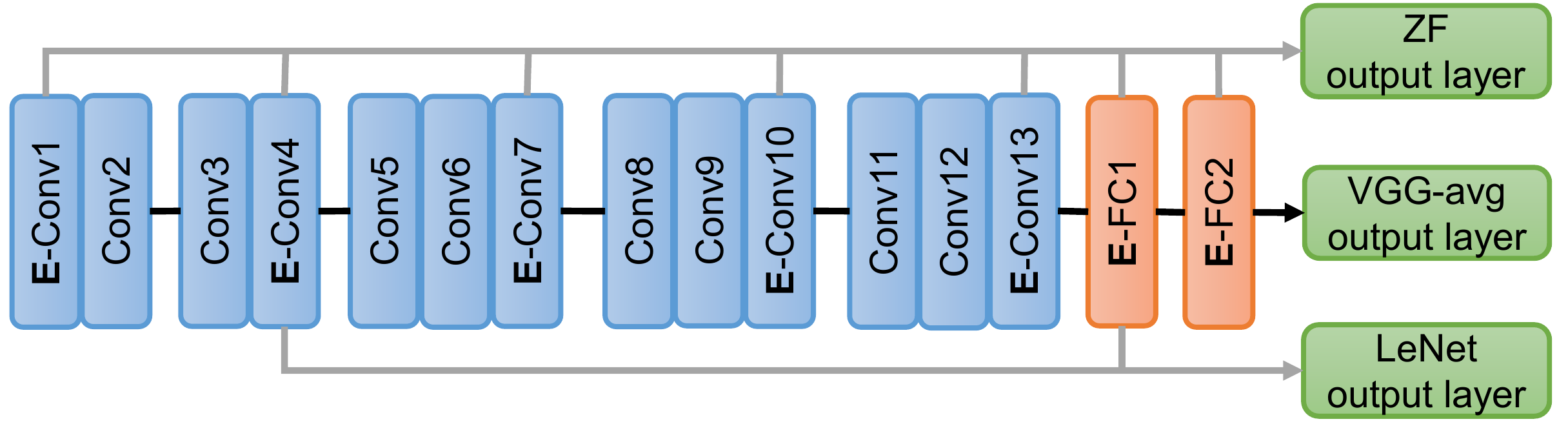}
		\end{center}
		\vspace{-0.35cm}
		\caption{Example of merging three models, ZF, VGG-avg, and LeNet into a single one for the inference stage via our NeuralMerger.}
		\label{fig:mergemore}
	\end{figure}

	\vspace{0.2cm}
	\noindent \textbf{Motivation of Our Study}:
	Merging existing neural networks has a great potential for real-world applications.
	%
	To tackle multiple recognition tasks in a single system based on either unique or various signal sources, a typical approach is to design a new model and train the model on the union datasets of these tasks, eg., \cite{DBLP:journals/corr/KaiserGSVPJU17,DBLP:journals/corr/AytarVT17}.
	Such ``learn-them-all" approaches train a single complex model to handle multiple tasks simultaneously.
	However, two issues may arise.
	First, it is hard to choose a suitable neural-net architecture for learning all the tasks well in advance;
	hence, a trial-and-error process is required to conduct suitable architectures.
	Second, learning from a random initial with large training data of different types could be demanding.
	To tackle these issues, the networks with bridging layers among the original models are conducted as a joint model for multi-task learning in \cite{RN274_multitask}.
	However, increased complexity and size of the joint model hinder their availability on resource-limited or edge devices in the inference stage.
	
	As many models trained for various tasks are available now, a practical way to integrate different functionalities in a system would be leveraging on these individual-task models.
	In this paper, we introduce an approach that merges different neural networks by removing the co-redundancy among their filters or weights.
	The proposed NeuralMerger can take advantage of existing well-trained models.
	Our approach merges them 
	via finding and sharing the representative codes of weights;
	the shared codes can still be refined by learning.
	To our knowledge, this is the first study on merging known-weights neural-nets into a more compact model. 
	Because our approach compresses the networks for weight sharing and redundancy removal, it is useful 
	for the deep-learning embedded system or edge computing in the inference stage.

	
	\vspace{0.15cm}
	\noindent \textbf{Overview of Our Approach}:
	When merging two different CNN models $C_A$ and $C_B$, the output is a CNN model consisting of jointly encoded convolution (\textbf{E}-Conv) and fully-connected (\textbf{E}-FC) layers.
	An overview of our approach is illustrated in Fig.~\ref{fig:concept} and an example of merging three models via our approach is given in Fig.~\ref{fig:mergemore}.
	
	Contributions of this paper are summarized as follows:
	
	\noindent (1) Given well-trained CNN models, the introduced NeuralMerger can merge them for multi-tasks even the models are different. The merging process preserves the general architectures of the well-trained networks and removes their redundancy. It avoids the cumbersome-design and trial-and-error process raised by the learn-them-all approaches.
	
	\noindent (2) The proposed method produces a more compact model to handle the original tasks simultaneously. The compact model consumes less computational time and storage than the compound model of the original networks. It has a great potential to be fitted in low-end systems.
	

	\section{Related Work}
	To simultaneously achieve various tasks via a single neural-net model, a typical way is to increase the output nodes (for multi-tasks) of a pre-chosen neural-net structure and train it from an initialization.
	In ~\cite{DBLP:journals/corr/KaiserGSVPJU17}, MultiModel architecture is introduced to allow input data to be images, sound waves, and text of different dimensions, and then converts them into a unified representation.
	In~\cite{DBLP:journals/corr/AytarVT17}, a deep CNN leverages massive synchronized data (sound and sentences paired with images) to learn an aligned representation.
	Nevertheless, as mentioned earlier, applying the learn-them-all approaches has to pay cumbersome training effort and intensive inference computation.
	
	Compressing a neural-net is an active direction to deploy the compact model on resource-limited embedded systems.
	To reduce the representation, binary weights and bit-wise operations are used in 
	\cite{hubara2016binarized} and 
	\cite{rastegari2016xnor}.
	Han et al.~\cite{Han16} introduce a three-stage pipeline: pruning redundant network connections, quantizing weights with a codebook, and Huffman encoding weights and index, to reduce the storage required by CNN. 
	Quantized CNN (Q-CNN)~\cite{Wu16} is proposed to address both the speed and compression issues, which splits the input layer space and applies vector quantization to each subspace.
	Researchers also try to prune filters and feature maps to directly reduce the computational cost~\cite{molchanov2016pruning}\cite{li2016pruning}\cite{he2017channel}.
	Transferring the learned knowledge from a large network to a small network is another important direction towards effective network compression.
	In~\cite{HintonDistilling14}, distilling the knowledge of an ensemble model into a smaller model is introduced. 
	
	
	Instead of compressing a single network, the goal of this study is to merge multiple networks simultaneously.
	Besides, our method can restore the performance of the jointly compressed models by fine-tuning it with the training samples.

	\section{Deep Model Integration}
	Assume that model A (or B) consists of $c_A$ (or $c_B$) convolution (Conv) followed by $f_A$ (or $f_B$) fully connected (FC) layers.
	Let $c_{min}=min(c_A,c_B)$.
	In our approach, a correspondence $(c_{A(i)},c_{B(i)})$ is established between the Conv layers for the alignment of the two models, $i \in \{1:c_{min}\}$; $A(\cdot)$ is a strictly increasing mapping from $\{1:c_{min}\}$ to $\{1:c_A\}$, and $B(\cdot)$ is a strictly increasing mappings from $\{1:c_{min}\}$ to $\{1:c_B\}$.
	Likewise, a correspondence $(f_{A(i)},f_{B(i)})$ is also established between the FC layers for $i \in \{1:f_{min}\}$.
	
	In our method, the merged layers have to be of the same type (Conv or FC).
	Given two layers, one in model A and the other in model B, the principle to merge them is finding a set of (fewer) exemplar codewords that represent the weights of the layers with small quantization errors.
	The layers are thus jointly compressed for redundancy removal.
	Below, we first consider unifying the Conv layers, and then the FC layers.
	
	\subsection{Merging Convolution Layers}
	
	Assume that some Conv layer in model A and some other in model B are to be merged.
	The layer of model A has the input volume size $N_A \times M_A \times d_A$, where $N_A\times M_A$ is the spatial size and $d_A$ is the depth (number of channels).
	The input volume is convolved with $p_A$ convolution kernels, where the size of each kernel is $n_A \times m_A \times d_A$.
	The output of the Conv layer in model A is thus a volume of $N_A \times M_A \times p_A$ (without loss of generality, assume that padded convolutions are used.)
	
	Likewise, similar notations apply to the respective layer in model B.
	An input volume of the size $N_B \times M_B \times d_B$ are convolved by $p_B$ convolution kernels of $n_B \times m_B \times d_B$.
	The output volume of that layer is of the size $N_B \times M_B \times p_B$.
	
	We aim to jointly encode the convolution coefficients.
	As there are $p_A$ (or $p_B$) convolution kernels in the layers of A (or B), we hope to find a new set of fewer (than $p_A+p_B$) exemplars to express the original ones so that the models are fused and the redundancy between them is removed.
	To this end, a viable way is to perform vector quantization (such as k-means clustering) on the convolution kernels and find a smaller number of codewords ($p<p_A+p_B$) to jointly represent the kernels compactly.
	However, it is demanding to make this method practicable because the kernel dimensions could be inconsistent (i.e., $n_A \neq m_A$ or $n_B \neq m_B$).
	
	To address this issue, we unify the different convolution kernels by using spatially  $1\times 1$ convolutions,
	so that merging CNNs with convolution kernels of different sizes is attainable.
	In the following, we review the operations in a convolution layer at first and then show how to separate the dimensions so that different layers are unified and jointly encoded.
	
	
	\subsubsection{Operations in Convolution Layer}
	The operations in a Conv layer of CNNs are reviewed as follows.
	Suppose $x\in R^{N\times M \times d}$ is the input volume (a.k.a.~3D tensor) to a Conv layer and $y\in R^{N\times M \times p}$ is the output volume.
	Assume that $p$ convolution kernels of size $n\times m \times d$ are applied to the layer, denoted as
	\begin{equation}
	\label{eq0}
	\{g^{(t)}\in R^{n\times m \times d} | t=1\cdots p\}.
	\end{equation}
	Then, the $t$-th channel output is obtained as $y_t=x \star g^{(t)}$, the volume convolution of $x$ and $g^{(t)}$,
	and the output $y$ is the concatenation of $y_t$,
	\begin{equation}
	\label{eq1.1}
	y =[y_{1}~y_{2}~\cdots~y_{p}].
	\end{equation}
	Let $x_u\in R^{N\times M}$ and $g_u^{(t)}\in R^{n\times m}$ respectively be the $u$-th channel of $x$ and $g^{(t)}$.  
	The volume convolution is formed by summing the 2D-convolution results of the $d$ channels:
	\begin{equation}
	\label{eq1.5}
	x\star g^{(t)}=\sum_{u=1}^{d} x_u\ast g_u^{(t)},
	\end{equation}
	where $\ast$ denotes the 2D convolution operator.
	
	In CNNs, various $n$ and $m$ (eg., $n=m=3, 5, 7 \cdots$) are used in existing networks.
	Particularly, when $n=m=1$, the volume convolution of size $1\times 1 \times d$ is often referred to as a $1\times 1$ convolution in CNNs for all $d$.
	
	\subsubsection{Kernel Decomposition in Spatial Directions}
	In the above, the volume convolution is computed as a spatially sliding operation (2D convolution) followed by a channel-wise summation along the depth direction.
	In this section, we show that, no matter what $n$ and $m$ are, it can be equivalently represented by $1\times 1$ convolutions via decomposing the kernel along with the spatial directions as follows.
	
	Given the kernel $g^{(t)}$, let $g^{(t)}_{[i_0,j_0],u}$ specify its entry at the spatial location ($i_0,j_0$) of the $u$-th channel. In particular, let $g_{[i_0,j_0]}^{(t)}\in R^d$ stand for the $1\times 1 \times d$ volume convolution at $(i_0,j_0)$; 
	e.g., a kernel of spatial size $5\times 5$ consists of 25 kernels of spatial size $1\times 1$, $(i_0, j_0) \in \{1 : 5\} \times \{1 : 5\}$, $\forall d$.
	
	Following the notation, we decompose an $n\times m \times d$ volume convolution into multiple $1\times 1 \times d$ convolutions and combine them with shift operators:
	Without loss of generality, we assume that the spatial sizes $n, m$ of the kernel are odd numbers and replace them with $w=(n-1)/2$ and $h=(m-1)/2$.
	The $t$-th channel output $y_t$ can be equivalently represented as
	\begin{equation}
	\label{eq2}
	y_t =\sum_{i_0=-w}^{w}\sum_{j_0=-h}^{h} S_{-i_0, -j_0}[x \star g_{[i_0,j_0]}^{(t)}],
	\end{equation}
	where $g_{[i_0,j_0]}^{(t)}$ ($-w\leq i_0 \leq w, -h\leq j_0 \leq h$) are the $1\times 1 \times d $ convolutions depicted above, and 
	$S$ is the shift operator, 
	\begin{equation}
	S_{i_0,j_0}[x](i,j,u)= x(i-i_0,j-j_0,u), \forall i, j, u,
	\end{equation}
	with $i,j$ the spatial location and $u$ $(1\leq u \leq d)$ the channel index.
	Hence, for all $n,m$, the $t$-th channel output of the volume convolution can be decomposed as the shifted sum of $nm$ $1\times 1$ convolutions via Eq.~\ref{eq2}.
	Then, the output $y$ is obtained via the concatenation in Eq.~\ref{eq1.1}.
	
	
	
	
	
	To address the issue caused by dimension mismatch in merging two convolution layers, we then propose to take the representation of $1 \times 1$ convolutions for both layers. 
	Hence, a kernel in model A is decomposed into $C_A=n_A m_A$ convolutions of size $1\times 1 \times d_A$ and that in model B is decomposed into $C_B=n_B m_B$ convolutions of size $1\times 1 \times d_B$.
	The kernel is then unified into $1 \times 1$ in the spatial domain no matter whether $n_A$ (or $m_A$) equals to $n_B$ (or $m_B$).
	
	\subsubsection{Kernel Separation along Depth Direction}
	Then, we seek to jointly express the $C_{AB}$ $1 \times 1$ convolutions by a compact representation so that the two layers are co-compressed, where $C_{AB}=p_A C_A+p_B C_B$ and $p_A, p_B$ are the numbers of kernels of the layers in A and B, respectively.
	Though the subspace dimensions in the spatial domain are consistent $(1\times 1)$ now, they are still inconsistent in the depth direction ($d_A$ vs.~$d_B$) and thus crucial to be jointly clustered.
	To address this problem, we simply separate the $1\times 1\times d$ kernel $g_{[i_0,j_0]}^{(t)}$ into non-overlapping $1\times 1\times r$ kernels along the depth direction ($r<d$).
	As the convolution in CNNs are summation-based in the depth direction, we divide the kernel $g_{[i_0,j_0]}^{(t)}\in R^d$ into $\lceil d/r \rceil$ vectors of dimension $r$,
	\begin{equation}
	\label{eq7}
	g_{[i_0,j_0];\langle v \rangle}^{(t)}\in R^r, v=1, \cdots, \lceil d/r \rceil,
	\end{equation}
	where $g_{[i_0,j_0];\langle v \rangle}^{(t)}$ (of size $1\times 1\times r$) is the $v$-th segment of the original kernel.
	The output $y_t$ in Eq.~\ref{eq2} then becomes
	\begin{equation}
	\label{eq8}
	y_t =\sum_{v=1}^{\lceil d/r \rceil}\sum_{i_0=-w}^{w}\sum_{j_0=-h}^{h} S_{-i_0, -j_0}[x_{\langle v \rangle} \star g_{[i_0,j_0];\langle v \rangle}^{(t)}],
	\end{equation}
	where $x_{\langle v \rangle} \in R^{N\times M \times r}$ is the $v$-th 
	sub-volume of the input $x$ for  $d=d_A$ or $d_B$.
	Specifically, a spatially $1\times 1$ kernel is respectively segmented into $\rho_A=\lceil d_A/r \rceil$ (or $\rho_B=\lceil d_B/r \rceil$) kernels of dimension $r$ in model A (or B), where the last segment is padded with zero if necessary.
	
	Let $\rho=min(\rho_A, \rho_B)$.
	There are then $C_{AB}$ kernels of size $1\times 1\times r$ for the segment $1\leq v \leq \rho$.
	To jointly represent the kernels of both layers, we use $C$ codewords ($C<C_{AB})$ in the dim-$r$ space to encode the convolution coefficients compactly.
	We run the k-means algorithm with various initials for the $C_{AB}$ vectors and then select the results yielding the least representation error to produce the $C$ codewords (i.e., cluster centers of k-means), for $v\in \{1,\cdots, \rho\}$.
	\footnote{For those remaining segments, $\rho+1\leq v \leq max(\rho_A,\rho_B)$, we also use $C$ codewords to encode the $p_AC_A$ (or $p_BC_B$) dim-$r$ vectors in the respective subspaces if $d_A>d_B$ (or $d_A<d_B$).}
	
	
	\begin{figure} [t]
		\begin{center}
			\includegraphics[width=0.95\columnwidth]{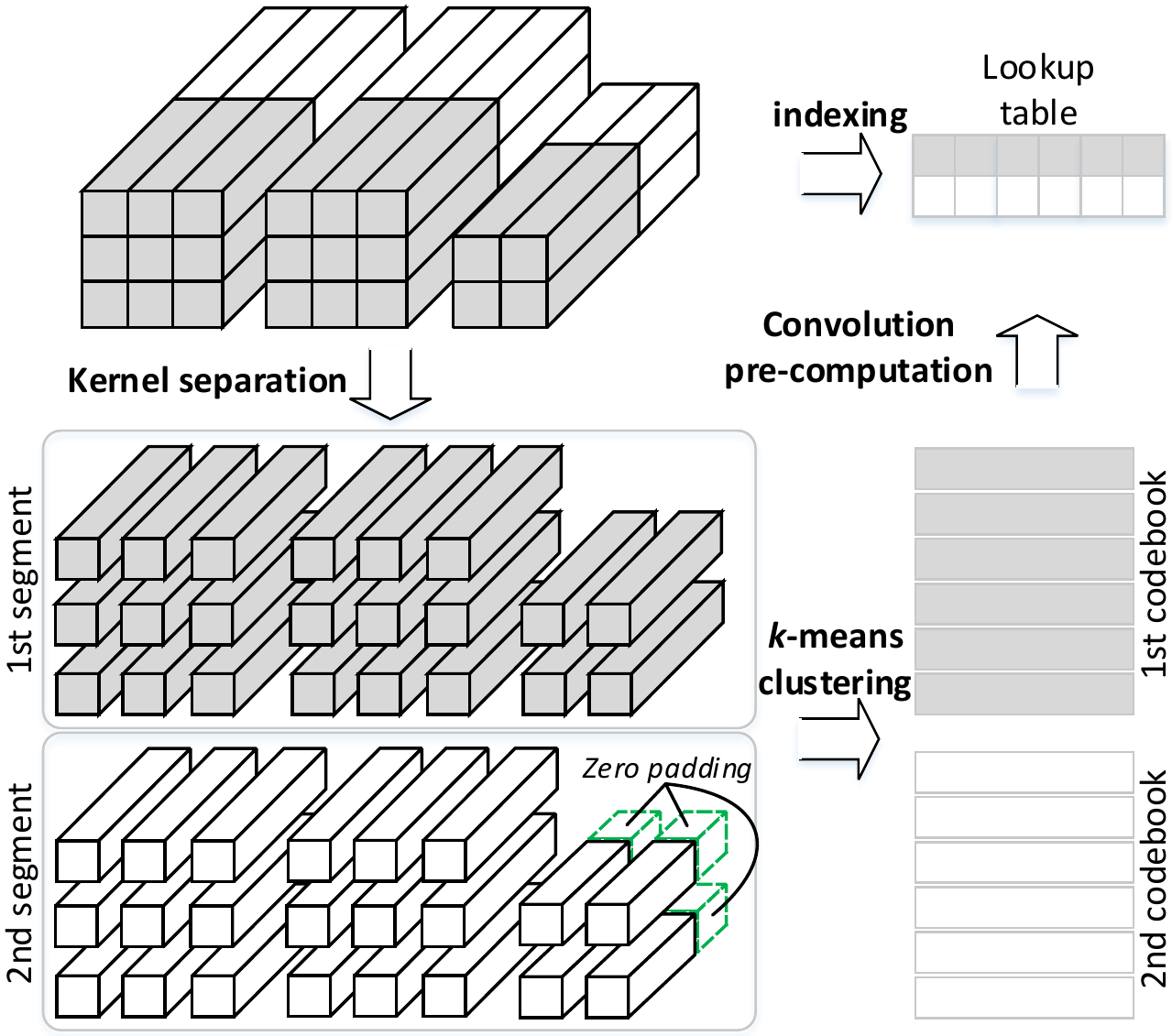}
		\end{center}
		\vspace{-0.3cm}
		\caption{Illustration of merging two models' Conv layers having the kernels of spatial size $3\times3$ and $2\times 2$, respectively; each layer is divided into 2 segments. They are decomposed into spatially $1\times 1$ kernels and the kernels in every segment is clustered via k-means clustering to build a coodbook. The convolutions are pre-computed on the codebook, and a lookup table is built to index the results.}
		\label{fig:decomposition}
	\end{figure}
	
	\subsubsection{E-Conv Layer and Weights Co-use}
	\label{sec:merge}
	The merged convolution layer (called the \textbf{E}-Conv layer), is a newly-formed layer where the weights are co-used among the convolution kernels:
	Denote the $C$ codewords in the $v$-th subspace to be $\{b_{c,v}\in R^r|c=1:C\}$.
	We then replace each dim-$r$ kernel at the spatial site $(i_0,j_0)$ in the subspace $v$ (namely, $g_{[i_0,j_0];\langle v \rangle}^{(t)}\in R^r$) with $b_{\pi(i_0,j_0,v,t);v}$, the closest codeword in the dim-$r$ space, where $\pi(i_0,j_0,v,t)\in\{1:C\}$ is the code-assignment mapping.
	Eq.~\ref{eq8} is then simplified as
	\begin{equation}
	\label{eq9}
	y_t =\sum_{v=1}^{\lceil d/r \rceil}\sum_{i_0=-w}^{w}\sum_{j_0=-h}^{h} S_{-i_0, -j_0}[x_{\langle v \rangle} \star b_{\pi(i_0,j_0,v,t);v}].
	\end{equation}
	Because the number of codewords $C$ is fewer than that of the total kernel vectors $C_{AB}$, Eq.~\ref{eq9} can be executed more efficiently via computing the $1\times1$ convolutions of the $C$ codewords at first:
	\begin{equation}
	\label{eq10}
	x_{\langle v \rangle} \star b_{c,v}~~~(\forall c=1 \cdots C, v=1 \cdots \rho),
	\end{equation}
	and then storing the results in a lookup table at run-time.
	The run-time operation of $1\times1$ convolution is thus replaced by table indexing.
	Hence, the convolution kernels of the two models A and B are representationally shared in a compact codebook, $\{b_{c,v}|v=1:C\}$, and the computation time is saved.
	An illustration of the \textbf{E}-Conv layer is given in Fig.~\ref{fig:decomposition}.

	When choosing the codewords $C$ fewer, the amount of convolution coefficients is reduced to $C/C_{AB}$.
	The merged model is thus co-compressed as it consumes less storage than the total required for the two convolution layers.
	As for the computational speed, each $x_{i,j;\langle v\rangle}\in R^r$ is replaced with an index and there are $NMd/r$ entries for indexing in $x_{\langle v \rangle}$, where $i,j$ are the spatial location.
	Let $\tau_x$ and be the time unit for a table-indexing operation and $\tau_r$ be the time unit for a $1\times 1\times r$ convolution.
	The speedup ratio is then $(C\tau_r+NMd\tau_x/r)/(C_{AB}\tau_r$) in terms of the complexity.
	Hence, when the codewords $C$ is fewer or the subspace dimension $r$ is larger, the speedup is getting higher.
	
	
	
	
	\subsubsection{Derivatives of the Merged Layer}
	Besides condensing and unifying the convolution operations, the \textbf{E}-Conv layer is also differentiable and so end-to-end back-propagation learning still remains realizable.
	However, evaluating the derivatives would be hard based on the table-lookup structure as the indices are not continuous.
	Hence, the table is used for the inference stage in our approach.
	While for learning, we slightly change the form of Eq.~\ref{eq9} to conduct the derivatives of $y_{i,j;t}$ (the output at the spatial location $i,j$ of channel $t$) to $\{b_{c,v}\}$ (the codewords).
	From Eq.~\ref{eq9}, 
	\begin{equation}
	\label{eq11}
	y_{i,j;t} =\sum_{v=1}^{\lceil d/r \rceil}\sum_{i_0=-w}^{w}\sum_{j_0=-h}^{h} \langle x_{i+i_0,j+j_0,\langle v \rangle}, b_{\pi(i_0,j_0,v,t);v} \rangle,
	\end{equation}
	where $\langle\cdot,\cdot\rangle$ is the inner product.
	Let $\Phi=[b_{c,v}]\in R^{r\times C}$ be the matrix whose columns are the dim-$r$ codewords of the $v$-th segment.
	Let $\beta_{i_0,j_0,v,t}\in R^C$ be the one-hot vector where the $c$-th entry in $\beta_{i_0,j_0,v,t}$ is $1$ if $c=\pi(i_0,j_0,v,t)$; otherwise the entry is $0$.
	Then, the codeword mapping $b_{\pi(i_0,j_0,v,t);v}$ in Eq.~\ref{eq9} can be replaced by $b_{\pi(i_0,j_0,v,t);v}=\Phi\beta_{i_0,j_0,v,t}$.
	Hence, the derivative of the \textbf{E}-Conv layer is conducted as
	\begin{equation}
	\label{eq12}
	\frac{\partial y_{i,j;t}}{\partial \Phi} = \sum_{v=1}^{\lceil d/r \rceil}\sum_{i_0=-w}^{w}\sum_{j_0=-h}^{h} x_{i+i_0,j+j_0,\langle v \rangle}\beta_{i_0,j_0,v,t}^T.
	\end{equation}
	As $\Phi$ is the matrix consisting of the codewords $\{b_{c,v}\}$, they can then be fine-tuned via the gradients for learning. 

	\subsection{Merging Fully-connected Layers}
	The volume input to an FC layer is re-shaped to a vector in general.
	Let $x_F \in R^{N_I}$ be the input vector of an FC layer and $y_F \in R^{N_O}$ be its output.
	Then $y_F=\mathbf{W}x_F$, where $\mathbf{W} \in R^{N_O\times N_I}$ are the weights of the FC layer.
	
	Unlike Conv layers that have sliding operations, all the operations in FC are summation-based.
	Thus, given the two weight matrices of models A and B, namely, $\mathbf{W}_A$ and $\mathbf{W}_B$, we simply divide them into length-$r$ segments along the row direction.
	The dim-$r$ weight vectors in the same segment are then clustered via k-means algorithm and $C$ codewords are found.
	In this way an \textbf{E}-FC layer is built as well, and
	It is easy to show that the \textbf{E}-FC layer is also differentiable. 
	
	\subsection{End-to-end Fine Tuning} 
	As both \textbf{E}-Conv and \textbf{E}-FC layers are differentiable, once their codebooks are constructed, we can then fine-tune the entire model from partial or all training data through end-to-end back-propagation learning.
	The training data are referred to as calibration data and the fine-tuning process is called the \emph{calibration training} in our work.
	Codebooks are also used for the weights quantization in the single-model compression approach~\cite{Wu16}.
	However, unlike our work that the codebooks in all layers are tunable in an end-to-end manner, only the codebook in a single layer can be tuned per time (with the other layers fixed) in~\cite{Wu16}, making its learning process inefficient and demanding to seek better solutions.
	Besides, our approach merges and jointly compresses multiple models, instead of only a single model.
	
	Two error terms are combined for the minimization in our calibration training.
	One is the classification (or regression) loss utilized in the original models A and B.
	The other is the layer-wise output mismatch error; 
	when applying the input $x_I$ to the model A (or B), the output of every layer in the merged model should be close to the output of the associated layer in A (or B), and $L_1$ norm is used to measure this error.
	We use a framework (TensorFlow~\cite{abadi2016tensorflow}) to implement the calibration training. 
	
	In the inference stage, to make the approach generalizable to edge devices that may not contain GPUs, we use CPU and OpenBLAS library~\cite{xianyi2012model,wang2013augem} to implement the merged model with the associated codebooks.
	To make fair comparisons, the Conv layers of the individual models compared with ours are realized via the unrolling convolution~\cite{chellapilla2006high,anwar2017structured} that converts the volume convolution into a single matrix product, which is commonly used as an efficient implementation for the Conv layer. Our code will be publicly available at GitHub\footnote{https://github.com/ivclab/NeuralMerger}.
	
	\section{Experiments}
	
	In this section, we present the results of several experiments conducted to verify the effectiveness of our approach.
	
	\subsection*{Merging Sound and Image Recognition CNNs}
	
	The first experiment is to merge two CNNs of heterogeneous signal sources: image and sound.
	Although the sources are different, the same network (LeNet~\cite{LeCun98}) is used.
	It is applied to the datasets of two tasks:
	
	\noindent \textbf{Sound20}\footnote{https://github.com/ivclab/Sound20}: This dataset contains 20 classes of sounds recorded from animal and instrument with 16636 training samples and 3727 testing samples. It is constructed using Animal Sound Data~\cite{keoghmonitoring} and Instrument Data~\cite{Juliani16}.
	The raw signals are converted to 2D spectrograms and then resized to $32\times32$ images for classification. 
	
	
	\noindent \textbf{Fashion-MNIST dataset}~\cite{xiao2017/online}: This dataset contains a collection of 60,000 training and 10,000 testing images of size $28\times28$ in 10 classes of dressing styles.
	
	\begin{table}[t]
		\centering
		\small
		\caption{The parameters of $r/C$ for 'ACCU' and 'LIGHT' settings in merging two LeNet models (while the classification layers are not co-compressed).}
		\label{Experimentone_param}
		\begin{tabular}{lccc}
			\toprule
			Para.       & Conv1       & Conv2      & Fc1           \\ \hline
			ACCU        & 1/64        & 8/128      & 8/128       \\
			LIGHT       & 1/64        & 32/128     & 8/64      \\
			\bottomrule
		\end{tabular}
	\end{table}

	\begin{table}[t]
		\centering
		\small
		\caption{Compression ratio, Speedup, and Accuracy drop of sound and image merged model. The speedup ratio is tested on the CPU of Intel(R) Xeon(R) CPU E5-2640 v4, in the single-thread model. 
		}
		\label{LeNetMerge}
		\begin{tabular}{lcccc}
			\toprule
			Para.   & Compr.      & Speedup                 & Sound $\downarrow$            & Image $\downarrow$  \\ \hline
			ACCU       & 10.4$\times$   & 1.3$\times$     & -0.06\%                   & 0.68\%      \\
			LIGHT       & 15.3$\times$   & 1.8$\times$    & 0.76\%                    & 1.40\%     \\
			\bottomrule
		\end{tabular}
	\end{table}

	LeNet consists of 2 Conv layers with 32 and 64 kernels of size $5 \times 5 \times 1$ and $5 \times 5 \times 32$, respectively, each followed by a $2\times 2$ max-pooling, and then an FC layer of 1024 units and a $\gamma$-way output ($\gamma$ is the number of classes).
	Before merged, LeNet can achieve 78.08\% and 91.57\% accuracy on the above sound (Sound20) and image recognition (Fashion-MNIST) tasks, respectively.
	The models are simply aligned and merged layer by layer since they are identical.
	
	There are two main parameters, $r$ and $C$, in our method, where $r$ is the length of the segment and $C$ is the number of codewords.
	As the input layer's depth of LeNet is 1, we set $r=1$ for the first Conv layer of both models and choose $C=64$ for it.
	Then, we alter the parameters of $r \in \{8, 16, 32\}$ and $C \in \{64, 128, 256\}$ for the remaining Conv- and FC-layers (except for the classification layer) and find the parameters per layer according to the performance when merging this layer only.
	By doing so, we conducted two settings, namely `ACCU' and `LIGHT', as shown in Table.~\ref{Experimentone_param}, where `ACCU' enforces the accuracy and `LIGHT' enforces the model size and inference speed, respectively.
	We test the inference speed of our model on 
	the Intel Xeon CPU E5-2640 v4 (single thread). 
	The C++ codes from~\cite{Wu16}, which adopts OpenBLAS library and realizes unrolling convolution based on Caffe's~\cite{jia2014caffe} CPU-mode implementation, is used and enhanced to realize the inference stage of our work.
	
\begin{figure} [t]
		\centering
		\includegraphics[width=2.5in]{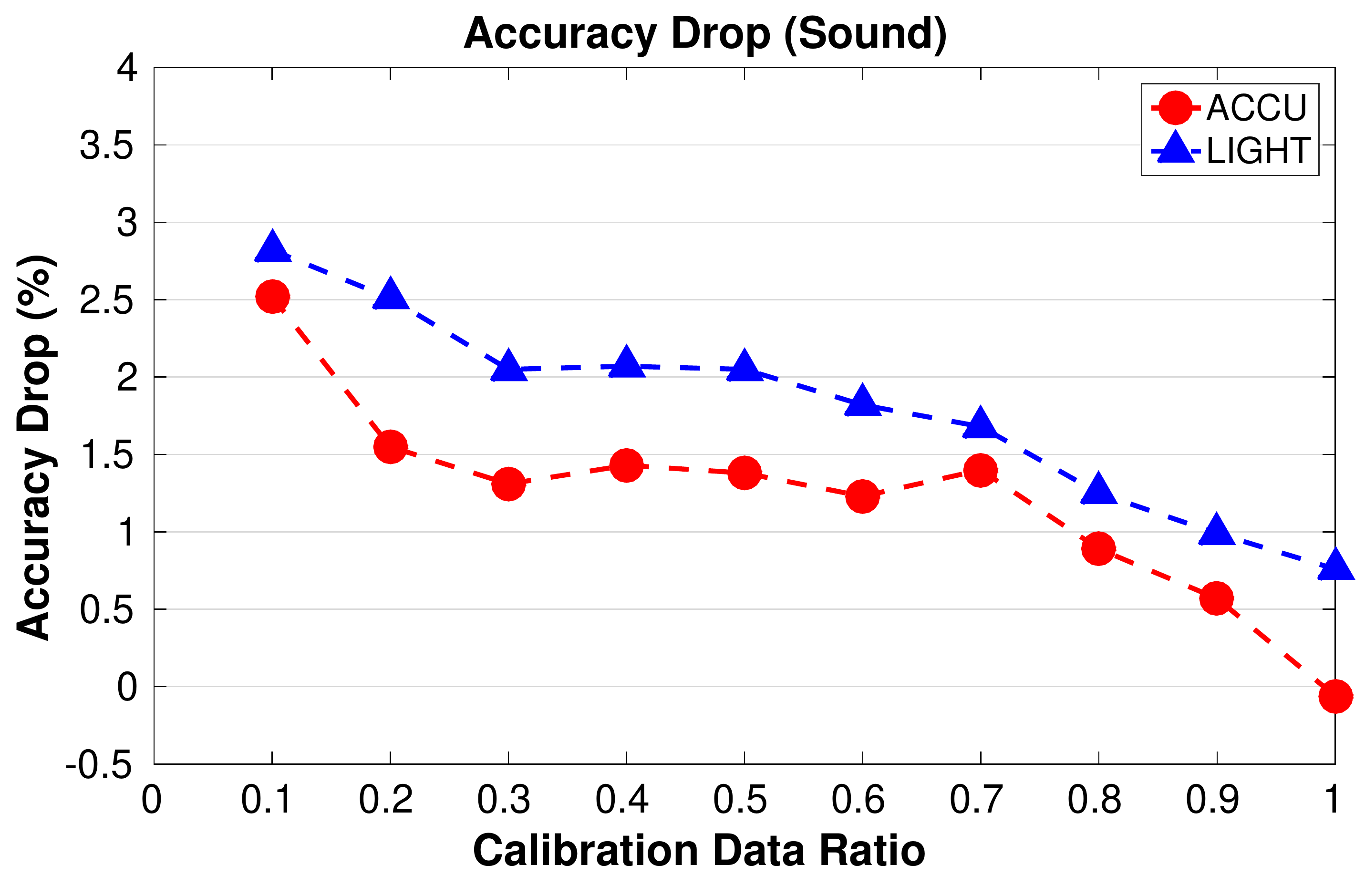}
		\includegraphics[width=2.5in]{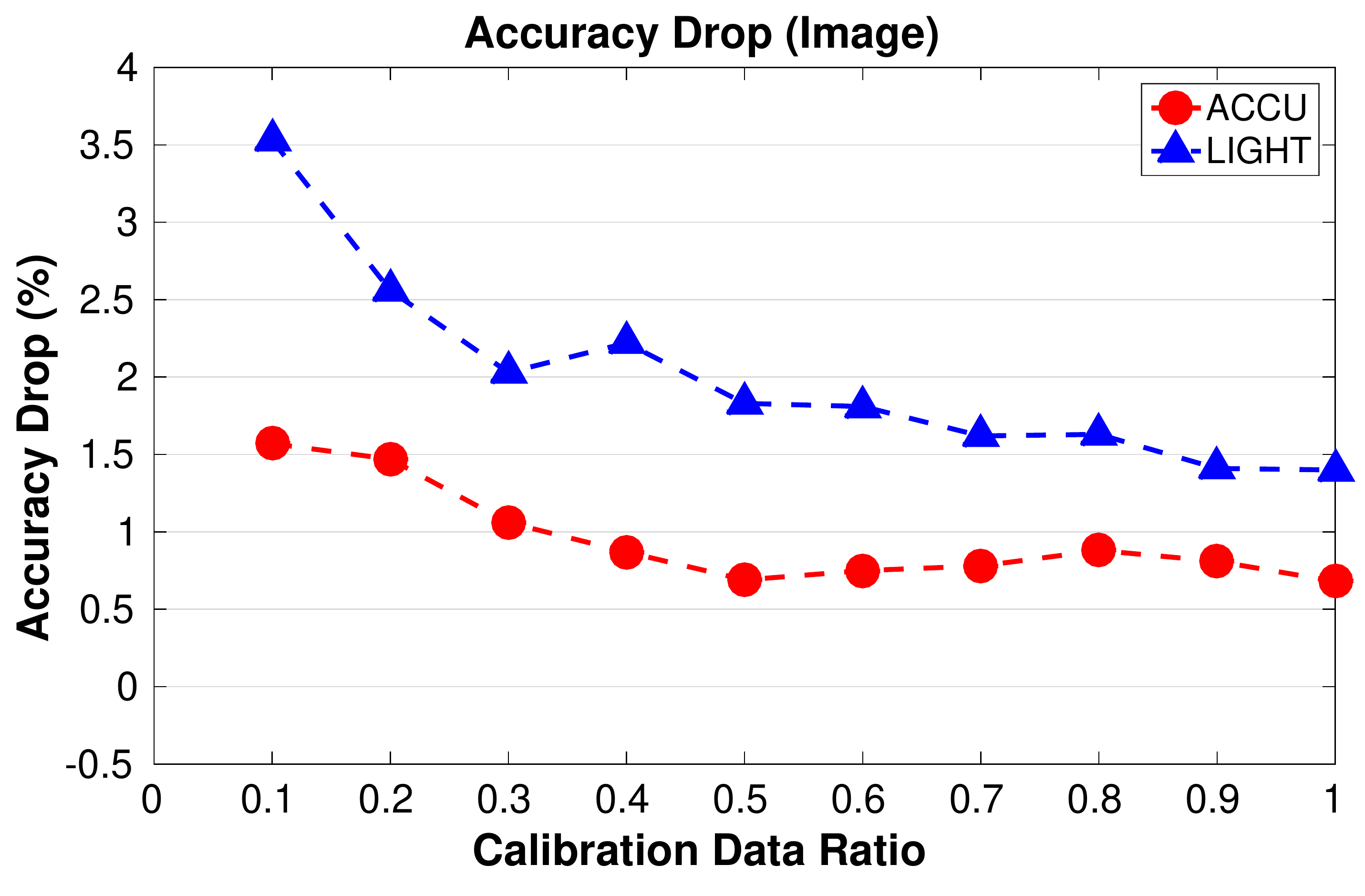}
		\caption{Accuracy drop vs. data ratio in calibration stage}	\label{fig:CalibrationDataRatio}
	\end{figure}

	The overall performance (after calibration training) and accuracy drop (denoted by $\downarrow$) are shown in Table~\ref{LeNetMerge}.
	With the setting of `ACCU', over 10 times of the compression ratio can be obtained on the joint model size, while only a negligible accuracy drop is imposed.
	As can be seen, for the image task, the accuracy of our merged model drops only $0.68\%$; for the sound task, the merged model even achieves a better accuracy than the original model ($-0.06\%$ accuracy drop).
	Under the `LIGHT' setting, our approach can achieve over 15 times of joint model-size compression
	with less than 1.1\% accuracy drop on average.
	Fig.~\ref{fig:CalibrationDataRatio} further shows the influence calibration data amount.
	It can be seen that no matter for the `ACCU' or `LIGHT' setting, more training data yield a lower accuracy drop in overall.

	\begin{table}[pt]
		\small
		\centering
		\caption{Comparison of the merged model (ACCU) with the individual sound and image models compressed via our approach. Each individual model and the merged model use the same parameters of $r/C$ and have identical model size and execution speed, but more tasks can be done in the merged model.}
		\label{LeNetMergeOrSingle}
		\begin{tabular}{lcccc}
			\toprule
			Parameter            & Image $\downarrow$            & Sound $\downarrow$       & Average Accuracy $\downarrow$      \\ \hline
			ACCU                &  0.68\%           & -0.06\%     & 0.31\%    \\ \hline
			Sound only           & N/A              & 0.40\%      & N/A                \\     
			Image only           & 0.51\%           & N/A         & N/A                \\
			\bottomrule
		\end{tabular}
	\end{table}

	As our approach finds several codewords per layer and then performs end-to-end calibration training to refine the codewords, it is also applicable to an individual model.
	One may wonder how the individual models perform when they are compressed via our approach.
	For an even comparison, we use the same $r/C$ settings to encode the single sound (or image) model and also fine-tune it using all training data.
	Hence, the compressed single model consumes the same resource as our merged one (i.e.,~they have exactly the same model size and execution time).
	As shown in Table~\ref{LeNetMergeOrSingle}, when the sound-only model is compressed, the accuracy is slightly dropped by 0.4\%; that is, our merged model can achieve even better accuracy while an additional functionality (image recognition) is added.
	On the other hand, when the image-only model is compressed, the accuracy-drop only changes little (from 0.51\% to 0.68\%) but an extra functionality (sound recognition) is supplemented.
	We owe that the two models have co-redundancy in between, and thus jointly encoding the models can take advantage of the additional inter-redundancy to achieve a better representation.

	\begin{table}[t]
		\centering
		\small
		\caption{The parameters of $r/C$ for 'ACCU' and 'LIGHT' settings in the clothing and gender merged model. Each group is a stack of two or three 3 $\times$ 3 Conv layers defined in VGG-16 model.  }
		\label{Experimenttwo_param}
		\begin{tabular}{lccccc}
			\toprule
			Para.       & Group1,2    & Group3   & Group 4,5  & Fc1    & Fc2             \\ \hline
			ACCU        & 32/256      & 16/256   & 8/256      & 4/128  & 4/64          \\
			LIGHT       & 32/128      & 32/128   & 16/128     & 8/128  & 8/64    \\
			\bottomrule
		\end{tabular}
	\end{table}

	\begin{table}[t]
		\centering
		\small
		\caption{Overall performance of the merged model for image classification of clothing and gender.}
		\label{TwoMerge}
		\begin{tabular}{lccccc}
			\toprule
			\multirow{2}{*}{Para.} & \multirow{2}{*}{Compr.} & \multicolumn{2}{c}{Clothing} & \multicolumn{2}{c}{Gender} \\
			&                              & Speedup      & Acc. $\downarrow$      & Speedup     & Acc. $\downarrow$     \\ \hline
			ACCU                      & 12.1$\times$                        & 1.5$\times$         & -0.83\%         & 1.8$\times$        & 0.27\%
			\\
			LIGHT                      & 20.1$\times$                        & 2.8$\times$         & 0.50\%         & 2.5$\times$         & 1.58\%          \\
			
			\bottomrule
		\end{tabular}
	\end{table}

	\begin{table}[pt]
		\small
		\centering
		\caption{Comparison of the merged model (ACCU setting) with the individual sound and image models compressed via our approach. Each individual model and the merged model use the same parameters of $r/C$ and have identical model size and execution speed, but more tasks can be done in the merged model.}
		\label{Ex2MergeOrSingle}
		\begin{tabular}{lcccc}
			\toprule
			Parameter            & Clothing  & Gender  & Average Accuracy $\downarrow$    \\ \hline
			ACCU                &  -0.83\%           & 0.27\%     & -0.28\%             \\ \hline
			Clothing only           & N/A              & 0.53\%         & N/A                          \\     
			Gender only           & -0.73\%           & N/A              & N/A                          \\
			\bottomrule
		\end{tabular}
	\end{table}

	\subsection*{Merging Clothing and Gender CNN Classifiers}
	In the second experiment, we merge two image classifiers (one for clothing and the other for gender recognition) into a single model holding double functionalities.
	The clothing classifier adopted is VGG-avg~\cite{yang2018supervised} which substitutes the first FC layer in the VGG-16 model~\cite{simonyan2014very} with average-pooling; it has been shown that VGG-avg achieves similar performance to VGG-16 while consuming far less storage.
	The gender classifier adopted is ZF-Net~\cite{ZFNet} that has fewer layers than VGG-avg.
	The overall structures and alignment between them can be found in Fig.~\ref{fig:mergemore}.
	
	The clothing and gender datasets employed are depicted as follows.
	The Adience Dataset~\cite{eidinger2014age} contains face pictures of different genders and ages, where 11,136 images (each resized to $227 \times 227$) are used for training and the other 3,000 are for testing.
	The Multi-View-Clothing dataset~\cite{DBLP:conf/mir/LiuCC16} contains clothing images of different views with over 250 attributes organized in a tree, where the frontal images (37,485 for training and 3,000 for testing) and the 13 attributes of the first two layers are used to construct a classifier of 13 classes.
	Before merged, ZF-Net and VGG-avg achieve 83.43\% and 89.80\% accuracies on the gender and clothing classification tasks, respectively.
	
	Because the kernel sizes in the two models are not always consistent, they are merged via the decomposition into $1\times 1 \times r$ convolutions as described in Section~\ref{sec:merge}.
	Similar to the process in the first experiment, we choose the parameters $r/C$ and conduct two settings `ACCU' and `LIGHT', where the former stresses accuracy and the latter emphasizes compression/speed. The detailed parameter settings are reported in Table~\ref{Experimenttwo_param}, and the overall performance is reported in Table~\ref{TwoMerge}.
	As the model sizes of both VGG-avg and ZF-Net are larger than that of LeNet, there are more redundancies between the models.
	Hence, an even better performance is achieved than that of experiment 1.
	In the `ACCU' setting, the compression ratio exceeds 12 times with even an increase of the average accuracy of the two tasks (negative average accuracy drop).
	The compression ratio is enhanced to more than 20 times with acceptable accuracy drops on the `LIGHT' setting.
	
	Besides, Table~\ref{Ex2MergeOrSingle} shows the results when compressing the individual models by using our approach under the same setting.
	It can be seen that similar (or even slightly better) accuracies can be obtained, with an additional merit that the merged model holds an extra functionality than the individual models.
	The results reveal that our approach can exploit both the intra- the inter-models redundancy and demonstrate satisfactory performance on merging deep learning models.

	\subsection*{Merging More Models}
	Finally, we show that our method can merge more models incrementally.
	The clothing-gender model in the above is added with another one, LeNet, for sound recognition (Fig.~\ref{fig:mergemore}).
	In Table~\ref{MoreMerge}, the accuracy-drop changes are negligibly small 
	but one more usage is added, which further confirms that our method can employ the redundancy among the models to enforce an efficient representation.
	

	\section{Conclusions}
	In this paper, we present a method that can unify multiple feed-forward models into a single but more compact one.
	To our knowledge, this is the first study on merging well-trained deep models for the inference stage. In our method, overall architectures of the original models are preserved.
	The unified model is still differentiable and can be fine-tuned to restore or enhance the performance. Experimental results show that the merged model can be extensively compressed under very limited accuracy drops, which makes our method potentially useful for deep model inference on low-end devices.
	

	\section*{Acknowledgement}
	This work was supported in part by the MOST under the grant MOST 107-2634-F-001-004.

	\begin{table}[t]
		\centering
		\small
		\caption{Accuracy Drop of merging three models(ACCU parameter)}
		\label{MoreMerge}
		\begin{tabular}{lcccc}
			\toprule
			Merged model             & Clothing $\downarrow$ & Gender $\downarrow$  & Sound $\downarrow$                \\ \hline
			Clothing + Gender     & -0.83\%    & 0.27\%   & N/A          \\
			Clothing + Gender + Sound    & -0.38\%    & 1.11\%   & 0.25\%     \\
			\bottomrule
		\end{tabular}
	\end{table}

	
	
	\bibliographystyle{named}
	\bibliography{ijcai18_conference}

\begin{thebibliography}{}

\bibitem[\protect\citeauthoryear{Abadi \bgroup \em et al.\egroup
  }{2016}]{abadi2016tensorflow}
Mart{\'\i}n Abadi, Ashish Agarwal, Paul Barham, Eugene Brevdo, Zhifeng Chen,
  Craig Citro, Greg~S Corrado, Andy Davis, Jeffrey Dean, Matthieu Devin, et~al.
\newblock Tensorflow: Large-scale machine learning on heterogeneous distributed
  systems.
\newblock {\em arXiv}, 2016.

\bibitem[\protect\citeauthoryear{Anwar \bgroup \em et al.\egroup
  }{2017}]{anwar2017structured}
Sajid Anwar, Kyuyeon Hwang, and Wonyong Sung.
\newblock Structured pruning of deep convolutional neural networks.
\newblock {\em ACM Journal on Emerging Technologies in Computing Systems
  (JETC)}, 13(3):32, 2017.

\bibitem[\protect\citeauthoryear{Aytar \bgroup \em et al.\egroup
  }{2017}]{DBLP:journals/corr/AytarVT17}
Yusuf Aytar, Carl Vondrick, and Antonio Torralba.
\newblock See, hear, and read: Deep aligned representations.
\newblock {\em CoRR}, abs/1706.00932, 2017.

\bibitem[\protect\citeauthoryear{Chellapilla \bgroup \em et al.\egroup
  }{2006}]{chellapilla2006high}
Kumar Chellapilla, Sidd Puri, and Patrice Simard.
\newblock High performance convolutional neural networks for document
  processing.
\newblock In {\em Tenth International Workshop on Frontiers in Handwriting
  Recognition}. Suvisoft, 2006.

\bibitem[\protect\citeauthoryear{Eidinger \bgroup \em et al.\egroup
  }{2014}]{eidinger2014age}
Eran Eidinger, Roee Enbar, and Tal Hassner.
\newblock Age and gender estimation of unfiltered faces.
\newblock {\em {IEEE} TIFS}, 9(12):2170--2179, 2014.

\bibitem[\protect\citeauthoryear{Han \bgroup \em et al.\egroup }{2016}]{Han16}
Song Han, Huizi Mao, and William~J Dally.
\newblock Deep compression: Compressing deep neural networks with pruning,
  trained quantization and huffman coding.
\newblock {\em ICLR}, 2016.

\bibitem[\protect\citeauthoryear{He \bgroup \em et al.\egroup
  }{2017}]{he2017channel}
Yihui He, Xiangyu Zhang, and Jian Sun.
\newblock Channel pruning for accelerating very deep neural networks.
\newblock In {\em International Conference on Computer Vision (ICCV)},
  volume~2, page~6, 2017.

\bibitem[\protect\citeauthoryear{Hinton \bgroup \em et al.\egroup
  }{2014}]{HintonDistilling14}
Geoffrey Hinton, Oriol Vinyals, and Jeff Dean.
\newblock Distilling the knowledge in a neural network.
\newblock {\em NIPS Workshops}, 2014.

\bibitem[\protect\citeauthoryear{Hubara \bgroup \em et al.\egroup
  }{2016}]{hubara2016binarized}
Itay Hubara, Matthieu Courbariaux, Daniel Soudry, Ran El-Yaniv, and Yoshua
  Bengio.
\newblock Binarized neural networks.
\newblock In {\em Advances in neural information processing systems}, pages
  4107--4115, 2016.

\bibitem[\protect\citeauthoryear{Jia \bgroup \em et al.\egroup
  }{2014}]{jia2014caffe}
Yangqing Jia, Evan Shelhamer, Jeff Donahue, Sergey Karayev, Jonathan Long, Ross
  Girshick, Sergio Guadarrama, and Trevor Darrell.
\newblock Caffe: Convolutional architecture for fast feature embedding.
\newblock {\em arXiv preprint arXiv:1408.5093}, 2014.

\bibitem[\protect\citeauthoryear{Juliani}{2016}]{Juliani16}
Arthur Juliani.
\newblock Recognizing sounds (a deep learning case study), 2016.

\bibitem[\protect\citeauthoryear{Kaiser \bgroup \em et al.\egroup
  }{2017}]{DBLP:journals/corr/KaiserGSVPJU17}
Lukasz Kaiser, Aidan~N. Gomez, Noam Shazeer, Ashish Vaswani, Niki Parmar, Llion
  Jones, and Jakob Uszkoreit.
\newblock One model to learn them all.
\newblock {\em CoRR}, abs/1706.05137, 2017.

\bibitem[\protect\citeauthoryear{Keogh}{2012}]{keoghmonitoring}
Yuan Hao Bilson Campana~Eamonn Keogh.
\newblock Monitoring and mining insect sounds in visual space.
\newblock 2012.

\bibitem[\protect\citeauthoryear{LeCun \bgroup \em et al.\egroup
  }{1998}]{LeCun98}
Yann LeCun, Léon Bottou, Yoshua Bengio, and Patrick Haffner.
\newblock Gradient-based learning applied to document recognition.
\newblock {\em Proceedings of the IEEE}, 86(11):2278--2324, 1998.

\bibitem[\protect\citeauthoryear{Li \bgroup \em et al.\egroup
  }{2016}]{li2016pruning}
Hao Li, Asim Kadav, Igor Durdanovic, Hanan Samet, and Hans~Peter Graf.
\newblock Pruning filters for efficient convnets.
\newblock {\em arXiv preprint arXiv:1608.08710}, 2016.

\bibitem[\protect\citeauthoryear{Liu \bgroup \em et al.\egroup
  }{2016}]{DBLP:conf/mir/LiuCC16}
Kuan{-}Hsien Liu, Ting{-}Yen Chen, and Chu{-}Song Chen.
\newblock Mvc: A dataset for view-invariant clothing retrieval and attribute
  prediction.
\newblock In {\em Proceedings of ACM on International Conference on Multimedia
  Retrieval}, pages 313--316, 2016.

\bibitem[\protect\citeauthoryear{Molchanov \bgroup \em et al.\egroup
  }{2016}]{molchanov2016pruning}
Pavlo Molchanov, Stephen Tyree, Tero Karras, Timo Aila, and Jan Kautz.
\newblock Pruning convolutional neural networks for resource efficient
  inference.
\newblock 2016.

\bibitem[\protect\citeauthoryear{Rastegari \bgroup \em et al.\egroup
  }{2016}]{rastegari2016xnor}
Mohammad Rastegari, Vicente Ordonez, Joseph Redmon, and Ali Farhadi.
\newblock Xnor-net: Imagenet classification using binary convolutional neural
  networks.
\newblock In {\em European Conference on Computer Vision}, pages 525--542.
  Springer, 2016.

\bibitem[\protect\citeauthoryear{Ruder}{2017}]{RN274_multitask}
Sebastian Ruder.
\newblock An overview of multi-task learning in deep neural networks.
\newblock {\em arXiv preprint arXiv:1706.05098}, 2017.

\bibitem[\protect\citeauthoryear{Simonyan and
  Zisserman}{2014}]{simonyan2014very}
Karen Simonyan and Andrew Zisserman.
\newblock Very deep convolutional networks for large-scale image recognition.
\newblock {\em arXiv preprint arXiv:1409.1556}, 2014.

\bibitem[\protect\citeauthoryear{Wang \bgroup \em et al.\egroup
  }{2013}]{wang2013augem}
Qian Wang, Xianyi Zhang, Yunquan Zhang, and Qing Yi.
\newblock Augem: automatically generate high performance dense linear algebra
  kernels on x86 cpus.
\newblock In {\em Proceedings of the International Conference on High
  Performance Computing, Networking, Storage and Analysis}, page~25. ACM, 2013.

\bibitem[\protect\citeauthoryear{Wu \bgroup \em et al.\egroup }{2016}]{Wu16}
Jiaxiang Wu, Cong Leng, Yuhang Wang, Qinghao Hu, and Jian Cheng.
\newblock Quantized convolutional neural networks for mobile devices.
\newblock In {\em Proceedings of the IEEE Conference on Computer Vision and
  Pattern Recognition}, pages 4820--4828, 2016.

\bibitem[\protect\citeauthoryear{Xiao \bgroup \em et al.\egroup
  }{2017}]{xiao2017/online}
Han Xiao, Kashif Rasul, and Roland Vollgraf.
\newblock Fashion-mnist: a novel image dataset for benchmarking machine
  learning algorithms, 2017.

\bibitem[\protect\citeauthoryear{Yang \bgroup \em et al.\egroup
  }{2018}]{yang2018supervised}
Huei-Fang Yang, Kevin Lin, and Chu-Song Chen.
\newblock Supervised learning of semantics-preserving hash via deep
  convolutional neural networks.
\newblock {\em IEEE transactions on pattern analysis and machine intelligence},
  40(2):437--451, 2018.

\bibitem[\protect\citeauthoryear{Zeiler and Fergus}{2014}]{ZFNet}
Matthew~D Zeiler and Rob Fergus.
\newblock Visualizing and understanding convolutional networks.
\newblock In {\em Proceedings of the European Conference on Computer Vision},
  pages 818--833, 2014.

\bibitem[\protect\citeauthoryear{Zhang \bgroup \em et al.\egroup
  }{2012}]{xianyi2012model}
Xianyi Zhang, Qian Wang, and Yunquan Zhang.
\newblock Model-driven level 3 blas performance optimization on loongson 3a
  processor.
\newblock In {\em IEEE International Conference on Parallel and Distributed
  Systems}, pages 684--691. IEEE, 2012.

\end{thebibliography}
	
\end{document}